\documentclass[sigconf, nonacm]{acmart}
\usepackage{hyperref}       
\usepackage{url}            
\usepackage{graphicx}
\usepackage{subfigure}
\usepackage[a-1b]{pdfx}

\usepackage{caption}
\usepackage{subfiles}
\usepackage{tabularx}
\usepackage{etoolbox}
\usepackage{bbm}
\usepackage{multirow}
\usepackage{enumitem}
\usepackage{titlesec}
\usepackage{amsmath}
\usepackage{amsthm}
\usepackage{xspace}
\usepackage{float}
\usepackage{booktabs}       
\usepackage{amsfonts}       
\usepackage{nicefrac}       
\usepackage{soul}
\usepackage{setspace}
\newcommand\vldbdoi{10.14778/3476311.3476402}
\newcommand\vldbpages{3178-3181}
\newcommand\vldbvolume{14}
\newcommand\vldbissue{12}
\newcommand\vldbyear{2021}
\newcommand\vldbauthors{\authors}
\newcommand\vldbtitle{\shorttitle} 
\newcommand\vldbavailabilityurl{http://vldb.org/pvldb/format_vol14.html}
\newcommand\vldbpagestyle{empty} 

\titlespacing{\paragraph}{%
  0pt}{
  0.3\baselineskip}{
  1em}

\begin{document}
\title{Managing ML Pipelines: Feature Stores and the Coming Wave of Embedding Ecosystems}
\author{Laurel Orr, Atindriyo Sanyal, Xiao Ling, Karan Goel, and Megan Leszczynski}
\affiliation{%
  \institution{Stanford University, Uber AI, Apple}
}
\email{{lorr1,kgoel,mleszczy}@cs.stanford.edu, atin@uber.com, xiaoling@apple.com}

\begin{abstract}
The industrial machine learning pipeline requires iterating on model features, training and deploying models, and monitoring deployed models at scale. Feature stores were developed to manage and standardize the engineer's workflow in this end-to-end pipeline, focusing on traditional tabular feature data. In recent years, however, model development has shifted towards using self-supervised pretrained embeddings as model features. Managing these embeddings and the downstream systems that use them introduces new challenges with respect to managing embedding training data, measuring embedding quality, and monitoring downstream models that use embeddings. These challenges are largely unaddressed in standard feature stores. Our goal in this tutorial is to introduce the feature store system and discuss the challenges and current solutions to managing these new embedding-centric pipelines.
\end{abstract}

\maketitle

\pagestyle{\vldbpagestyle}
\begingroup\small\noindent\raggedright\textbf{PVLDB Reference Format:}\\
\vldbauthors. \vldbtitle. PVLDB, \vldbvolume(\vldbissue): \vldbpages, \vldbyear.\\
\href{https://doi.org/\vldbdoi}{doi:\vldbdoi}
\endgroup
\begingroup
\renewcommand\thefootnote{}\footnote{\noindent
This work is licensed under the Creative Commons BY-NC-ND 4.0 International License. Visit \url{https://creativecommons.org/licenses/by-nc-nd/4.0/} to view a copy of this license. For any use beyond those covered by this license, obtain permission by emailing \href{mailto:info@vldb.org}{info@vldb.org}. Copyright is held by the owner/author(s). Publication rights licensed to the VLDB Endowment. \\
\raggedright Proceedings of the VLDB Endowment, Vol. \vldbvolume, No. \vldbissue\ %
ISSN 2150-8097. \\
\href{https://doi.org/\vldbdoi}{doi:\vldbdoi} \\
}\addtocounter{footnote}{-1}\endgroup


\section{Introduction}
\label{sec:introduction_1}

Building industrial machine learning pipelines is an iterative cycle of gathering and curating training data, training and deploying models, and monitoring the model once put into production. When errors or undesirable behavior are found, the cycle is repeated. Without tools to manage this process, production models become hard to maintain and difficult to reproduce. In 2017, a subset of the authors built the first industrial {\em feature store}~\cite{michaelangelo}---a system designed to standardize and manage model features and workflows.
The feature store both reduced engineer effort and improved model quality. Feature stores, however, have yet to adapt to a growing trend in model development: incorporating pretrained embeddings.

Pretrained embeddings are becoming standard inputs to modern machine learning pipelines. These embeddings are typically trained in a self-supervised fashion over massive data sets and encode knowledge about words, entities, graphs, and images. Once trained, they can provide lift in numerous downstream tasks like recommendation systems~\cite{naumov2019deep}, information retrieval~\cite{khattab2020colbert}, and data integration~\cite{mudgal2018deep}. 
A subset of the authors saw first-hand from their experience building and deploying an industrial self-supervised entity disambiguation system that pretrained embeddings are shifting industrial pipelines towards ``hands-free'' models that require limited hand-curated data and model engineering~\cite{overton2020re, karpathy20, molino2019ludwig}. There is therefore an increasing need for the next generation of feature store systems to help manage and monitor the embedding training data, pretrained embeddings, and downstream systems that consume the embeddings. 

\paragraph{Goal} The goal of this tutorial is to expose the interplay between data management and modern, self-supervised embedding ecosystems. We will first introduce feature store systems and the challenges they address. We will then introduce self-supervised pretrained embeddings and explain the new challenges associated with managing these embedding pipelines. We then explore how data management can help build, monitor, and maintain these self-supervised embedding ecosystems. Lastly, we will discuss future data management challenges and research directions.

\paragraph{Scope} This tutorial focuses on modern feature stores and the challenges with supporting embeddings as first class citizens in feature stores. We highlight that managing these self-supervised embedding systems {\em is} a fundamental data management problem.
This tutorial is intended for researchers with some familiarity with deep learning and pretrained embeddings who are interested in the interaction of data management and deep learning pipelines.

\begin{figure*}[ht!]
    \centering
    \includegraphics[width=0.75\textwidth]{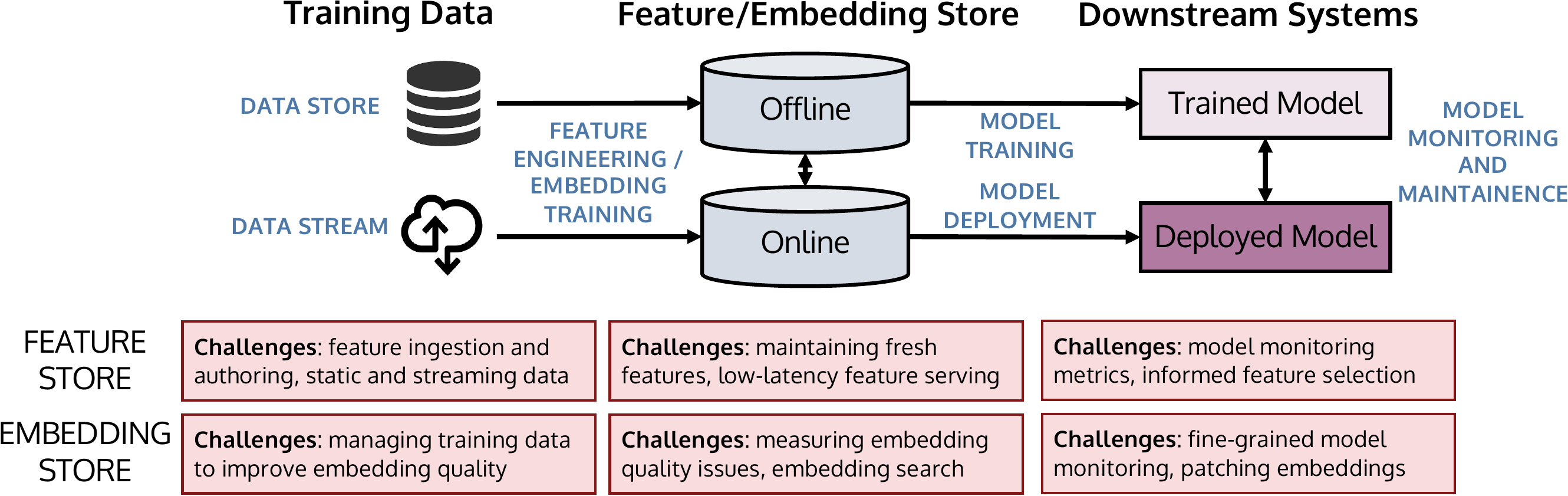}
    \caption{Modern machine learning pipeline with the challenges associated with managing a feature store (top row) and the {\em additional} challenges with managing an embedding ecosystem (bottom row).}
    \label{fig:ecosystem}
    \vspace{-0.2cm}
\end{figure*}

\paragraph{Outline} This tutorial is split into three parts over 1.5 hours.\footnote{This tutorial has not been presented in any venues prior and is most similar to recent SIGMOD and VLDB tutorials in data integration and cleaning with machine learning.}
\begin{enumerate}
    \item \textbf{Feature Stores}. We will give an overview of the modern machine learning pipeline and feature store systems. We will describe the core challenges these systems solve and give an overview of the technical contributions.
    \item \textbf{Embedding Ecosystems}. We will introduce pretrained embeddings and discuss the new challenges faced by feature stores in treating embeddings as first class citizens. We then discuss recent solutions to some of these challenges.
    \item \textbf{Future Directions and Challenges}. We will conclude with a discussion of the future directions and challenges.
\end{enumerate}

\section{ML Feature Stores}
\label{sec:feat_store_2}
In 2017, a subset of the authors built the first industrial feature store~\cite{michaelangelo}. Using this first-hand experience and the lessons learned, we introduce the modern ML pipeline and describe the challenges faced in maintaining and deploying models. We then introduce feature store systems and the technical innovations that help solve the aforementioned challenges. We focus the first part of this tutorial on traditional {\em tabular} feature data (i.e., not embeddings).

\subsection{Machine Learning Pipelines}
In industrial machine learning (ML) pipelines, as shown in \autoref{fig:ecosystem}, engineers need to quickly ingest training data for feature curation, train and deploy models, and monitor and maintain the model once deployed. We describe each step below.

\begin{enumerate}
    \item \textbf{Training Data}. Data is scraped, mined, or retrieved from a variety of different sources. The data needs to be cleaned, checked, and featurized for downstream models. 
    
    \textit{Challenges:} Engineers author custom features that, if not shared and managed, can result in repeated work and lack of definitional consistency. Feature definitions can become stale if not kept up-to-date as data changes over time.
    
    \item \textbf{Model Training and Deployment}. Using a set of features, engineers need to train and deploy models.
    
    \textit{Challenges:} As data changes over time and updates occur at different intervals, models can become stale if not given the most up-to-date features. Further, model reproducibility becomes a challenge as engineers try to keep up with changing data and model parameters.
    
    \item \textbf{Model Maintenance and Monitoring}. Once deployed, models need to be monitored and maintained.
    
    \textit{Challenges:} Models can struggle in the face of distribution shift and out-of-domain inputs~\cite{schelter2018challenges}. Further, once model errors are detected, engineers are often lacking guidance as to what features need to be corrected.
    
\end{enumerate}

\subsection{Feature Stores}
{\em Feature stores} (FSs) arose to address these challenges by providing a centralized repository of reusable features across the ML pipeline and automating the management of this pipeline~\cite{featurestore, michaelangelo, hopsworks, feast}.
Below, we dive into how feature stores address the three challenges above.

\subsubsection{Training Data}\mbox{}\\
Structured data can be in the form of raw tables as well as streams that users access when curating features. To facilitate sharing of features across an organization and maintaining features if they get updated, feature stores allow for {\em feature authoring and publishing}~\cite{alkowaileet2018end}. Users provide simple definitional metadata, e.g., the feature update cadence and a definition SQL query, and upload the definition to the FS. When the underlying data changes, the FS orchestrates the updates to the features based on the user-defined cadence.

For streaming features, users provide aggregation functions that are applied on the raw streaming features. The aggregated features are persisted to the online store and logged to the offline store.





\subsubsection{Model Training and Deployment}
\paragraph{\textbf{Feature Storage}}
Once features are curated, users need the ability to construct feature sets on the most recent data to train and deploy models. FSs support this workflow by partitioning features on date and providing APIs to allow for time based joins. Further, FSs must support feature quality metrics to support the detection and mitigation of feature errors. For example, FSs measure feature freshness, null counts, and mutual information across features. 

\paragraph{\textbf{Model Storage}}
Once a model is trained, relevant parameters and artifacts need to be stored for provenance and reproducibility. Although model storage is not traditionally part of a FS, some FSs~\cite{michaelangelo, hopsworks} do support model management by integrating a separate model store~\cite{vartak2016modeldb, gharibi2019modelkb}.

\paragraph{\textbf{Online Feature Serving}}
Once a model is deployed, features need to be continuously provided to deployed models even as the feature data is updated over time. To provide low latency feature serving, FSs are typically a dual datastore: one for offline training (e.g., SQL warehouse) and for online serving (e.g., in-memory DBMS). 

\subsubsection{Model Monitoring and Maintenance}\mbox{}\\
FSs must additionally support model quality metrics (as well as feature quality metrics described above)~\cite{vartak2016modeldb}. For example, FSs support critical model metrics such as training-deployment data skew and near real-time outlier and input drift detection. These metrics allow users to be informed of potential `gremlins' in the system. Once an error is discovered, engineers can use the FS metrics to detect the offending set of features and select a more optimal feature set for serving (or retraining).




\vspace{-0.2cm}
\section{Data Management and Embedding Ecosystems}
\label{sec:db_for_ml_3}
Drawing on our first-hand experiences developing entity embeddings across numerous downstream products at a large technology company, we now introduce self-supervised pretrained embeddings and the ecosystems around them. We then discuss new challenges and solutions with managing these ecosystems. We believe treating embeddings as first class citizens is the next evolution of feature stores.

\subsection{Embedding Ecosystem}

We define the embedding ecosystem as the embedding training data, embeddings, and downstream systems that consume them. As shown in \autoref{fig:ecosystem}, the embedding ecosystem pipeline is similar to that of the feature store. However, FSs are unable to support end-to-end embedding management. With embeddings, standard metrics and tools for managing tabular features are no longer adequate as embeddings are derived data. For example, embeddings are often compared by dot product similarity, and existing FS metrics such as null value count do not capture drifts or changes in embeddings with respect to this metric.

We now highlight the additional challenges associated with each step of the ML pipeline when incorporating embeddings (the same challenges from FSs still apply to embedding ecosystems).
\begin{enumerate}
    \item \textbf{Training Data:} In an embedding ecosystem, the raw training data is used to pre-train the embeddings. As this data is self-supervised, it will not be hand-labeled or curated.
    
    \textit{Challenges:} Embeddings will encode any inherent biases that exist in the self-supervised training data, e.g., the embeddings do not well represent rare things~\cite{orr2021bootleg, schick2020rare}.
    
    \item \textbf{Model Training and Deployment:} Once embeddings are trained, they need to be stored and served to downstream systems that use embeddings for training and deployment. 
    
    \textit{Challenges:} As embeddings get retrained and updated, just like features, the downstream models can become stale and out-of-date. Users need to understand and monitor the embedding changes and search over possible embeddings and select the best ones for their task. Unlike features, standard tabular metrics are inadequate for embeddings.
    
    \item \textbf{Model Maintenance and Monitoring:} Like with FSs, deployed models need to be monitored and maintained, especially with respect to the embedding inputs.
    
    \textit{Challenges:} Any inherent embedding quality issue will impact all downstream models using those embeddings. Users need to be able to understand and isolate downstream quality issues in the underlying embeddings. Once found, users need methods for correcting errors in downstream products.
\end{enumerate}

\subsubsection{Self-Supervised Training Data}\mbox{}\\
Unlike feature curation data, which is tabular and pre-labeled, self-supervised training data is often unstructured and is not hand-curated. This lack of manual curation results in data that may not accurately represent the data seen upon deployment \cite{bernstein2012direct, googlequeries, koh2020wilds} and is often biased toward popular things \cite{orr2021bootleg}. Embeddings trained on this data can inherit these biases. 

To improve embedding quality of rare things through training data management, recent work from \cite{orr2021bootleg} explored incorporating structured data into entity embedding pretraining through named entity disambiguation, the task of mapping from strings to things in a knowledge base. They showed that by adding structured data of the type of an entity and its knowledge graph relations, they could boost performance over rare entities by 40 F1 points. We believe merging structured and unstructured data is a promising management technique for improving quality in training data.

\subsubsection{Embedding Management}\mbox{}\\
Both traditional features and pretrained embeddings are served to downstream models. The uniqueness of an embedding ecosystem is that users need to be able to understand the difference in embedding quality as embeddings are updated over time and need guidance over what embeddings to use for their task.

To measure quality, \citet{wendlandt2018factors} and \citet{hellrich2016bad} discuss analyzing word embeddings with respect to an embedding's nearest neighbors. The work of \citet{leszczynski2020understanding} is uniquely looking at the quality of an embedding with respect to a downstream task. The authors define the metric of downstream instability, the number of predictions that change with different embeddings, to measure downstream embedding instability.
%

There is little available work on finding the right embedding to use, especially given compute or memory constraints. 
The work of \citet{may2019downstream} takes a first step by a variant of the eigenspace overlap score as a way of predicting downstream performance. However, their work is focused on measuring the performance of non-contextualized word embeddings.



\subsubsection{Fine-Grained Monitoring and Patching}\mbox{}\\
Downstream models in traditional FSs and embedding ecosystems need to be monitored and maintained. In an embedding ecosystem, however, the challenge is in giving users the tools to find meaningful subpopulations of errors and connecting the downstream errors to embedding quality issues. These errors then need to be corrected {\em through the underlying embedding}.

In terms of monitoring downstream models, recent works provide toolkits for measuring language model performance at a semantic, fine-grained level~\cite{goel2021robustness, ribeiro2020beyond}. \citet{goel2021robustness} in particular focuses on allowing users to define custom sub-population functions to explore performance across different models. 

Once an error is discovered, the challenge is in how to correct that error in the underlying embedding. By correcting the error in the embedding, all downstream systems using those embeddings will be patched, which maintains product consistency. The work in \citet{orr2021bootleg} gives a proof-of-concept that using data management techniques such as augmentation~\cite{chepurko2020arda}, weak supervision~\cite{ratner2017snorkel}, and slice-based learning~\cite{chen2019slice} can correct underperforming sub-populations of data~\cite{orr2021bootleg}.


\vspace{-0.2cm}
\section{Future Directions}
\label{sec:future_4}
We end with a discussion of future directions.

\paragraph{Embedding Enhanced Feature Stores} We believe the next evolution of a feature store is one with native support for embeddings. While we discussed some challenges and potential solutions, this is just the beginning. Users need tools for searching and querying these embeddings as well as support for versioning, provenance, and downstream quality metrics. For example, if an embedding gets updated but a model that uses it does not, the dot product of the embedding with model parameters can lose meaning which leads to incorrect model predictions. Further, performing these operations at industrial scale will be non-trivial as the size of embeddings and their associated models are continuing to increase. 

\paragraph{End-to-End Model Patching Through Data}
An open area of research is in automatically correcting the errors discovered in the downstream model error analysis through the underlying embedding. While prior work showed you can patch errors through methods like data augmentation and slice finding~\cite{orr2021bootleg, goel2020model}, there are remaining challenges in how to automate and manage this process. How can you predict if an augmentation strategy will have the desired result? If an embedding gets patched, what is the optimal way to propagate that patch downstream?

\section{Biographical Sketches}
\label{sec:bio_5}
\noindent\textbf{Laurel Orr} is a PostDoc in the Computer Science Department at Stanford University advised by Christopher R\'{e}. She graduated from the University of Washington in the Database group and is a lead on the Bootleg project, a self-supervised system for entity disambiguation. Bootleg is in production at Apple and used by academic research groups. She was awarded the NSF GRFP as a graduate student and is a current IC Postdoc Fellow.

\noindent\textbf{Atindriyo Sanyal} is a technical lead on the Michelangelo team at Uber AI. He leads various feature engineering efforts across Uber. Prior to that, he worked at LinkedIn and Apple where he was a Senior Software Engineer on Siri and was the founding engineer behind SiriKit (Siri API). He did his Masters at UCLA, where he worked at the Networks Research Lab building routing algorithms for pedestrians with skin conditions on open source navigation systems. He's one of the winners of Microsoft's Imagine Cup, won many hackathons at university, an IEEE presidential Award nominee, and a Math Olympiad winner.

\noindent\textbf{Xiao Ling} is a Machine Learning engineer at Apple where his work spans from information extraction for knowledge base construction to open-domain question answering. He earned his PhD in Computer Science and Engineering from the University of Washington in 2015. He was an early engineer at Lattice Data Inc., which was acquired by Apple in 2017.

\noindent\textbf{Megan Leszczynski} is a PhD student in the Computer Science Department at Stanford University advised by Christopher R\'{e}. She is one of the original developers of Bootleg, a self-supervised system for named entity disambiguation, which has since been deployed in industry. She has also given an invited lecture to the Stanford CS224N (NLP with Deep Learning) course led by Christopher Manning. Her research has been recognized with a NSF GRFP.

\noindent\textbf{Karan Goel} is a 3rd year CS PhD student at the Stanford AI Lab. He leads the Robustness Gym (RG) project, whose goal is to facilitate fine-grained evaluation and maintenance of ML models. RG is actively deployed at Salesforce with users in academia and industry. He wrote one of the first papers on "model patching", and his work has been recognized with a Siebel Scholarship (2018) and a Salesforce Research Grant (2020).

\begin{footnotesize}
\noindent \textbf{Acknowledgements} We acknowledge the support of the IC Postdoctoral Research Fellowship Program and NSF Graduate Research Fellowship under No. DGE-1656518.
\end{footnotesize}


\vspace{-0.2cm}
\bibliographystyle{ACM-Reference-Format}
\bibliography{references}

\end{document}